\documentclass[12pt,a4paper]{article}
\usepackage{amsmath}
\usepackage{amsfonts}
\usepackage{amssymb}
\usepackage{natbib}
\usepackage{algorithm}
\usepackage{algpseudocode}
\usepackage{amsthm}
\usepackage{dsfont}

\usepackage{graphicx}
\usepackage{mathtools}

\newtheorem{definition}{Definition}

\numberwithin{equation}{section}

\begin{document}

\begin{titlepage}
   \vspace*{\stretch{1.0}}
   \begin{center}
      \Large\textbf{A Bayesian Approach To Graph Partitioning}\\
      
      \Large\textbf{(April 24th, 2022)}  \\

	  \large\textit{Farshad Noravesh}\footnote{Email: noraveshfarshad@gmail.com}

   \end{center}
   \vspace*{\stretch{2.0}}
\begin{abstract}
A new algorithm based on bayesian inference for learning local graph conductance   based on Gaussian Process(GP) is given that uses advanced MCMC convergence ideas to create a scalable and fast algorithm for convergence to stationary distribution which is provided to learn the bahavior of conductance when traversing the indirected weighted graph. First metric embedding is used to represent the vertices of the graph. Then, uniform induced conductance is calculated for training points. Finally, in the learning step, a gaussian process is used to approximate the uniform induced conductance. MCMC is used to measure uncertainty of estimated hyper-parameters. 
\end{abstract}

\end{titlepage}

\section{Introduction}
There are many paradigms for graph partitioning and \citep{Buluc2016} covers some of them. In fact, there are many communities working on it from a different perspective. Graph partitioning is often done by computer scientists using linear programming(LP) and semi-definite programming(SDP) as is explained in \citep{Arora2009} in the framework of approximation algorithms.

Probabilists formulate the problem as follows:
\begin{definition}
Graph partitioning is decomposing the graph into a set of clusters $V_{i}$ such that the mixing rate inside the clusters is as much as possible while there are minimum number of edges connecting these clusters. 
\end{definition}
There are many proxies to control mixing rate of a markov chain such as conductance which is inspired by the seminal work of \citep{Jerrum1988}. Finding the conductance for all subsets is not a scalable task and grows exponentially. The novelty of the present paper is learning the qualitative behavior of conductance by variation, and observing what happens to the conductance when the distance to some reference vertices varies.
\citep{Orecchia2013} solves graph partitioning by solving some max flow computations. Other methods are based on finding sparse cuts like \citep{Noravesh2022} ,\citep{Andersen2009}.

\subsection{Sparse GP}
Gaussian processes have some known limitations if it is implemented classically. The following problems are just some major issues: \\
1- Tractability: The classical algorithm for GP is only tractable for gaussian observations.
$p(z|y,\rho,\Phi) \approx \prod_{i=1}^{N}  p(y_n|z_{n},\rho)N(z|0,c_{z,z}) $   \\
2- Choice of kernel and its hyperparameters. \\
3- Space complexity of $O(N^3)$ for inverting large matrices. So storing them takes huge space.\\
4- Posterior distribution describes a highly correlated high-dimensional variable. Thus simple MCMC sampling methods such as Gibbs sampling could be inefficient. \\
A variational approximation to the posterior which is sparse can answer all the above three issues simultaneously as is done in \citep{Hensman2015} and minimizes KL divergence in the variational inference framework. In fact,
\citep{Hensman2015},\citep{Snelson2005} reduced the computational complexity to $O(NM^2)$ (M is number of inducing points) by variational inducing point methodology which combines MCMC with variational inference. Following the idea of  \citep{Hensman2015}, many researchers such as \citep{Gomez2021} began to explore different ways to utilize the effectiveness of MCMC for gaussian processes since  MCMC framework could handle any general factorized or nonfactorized likelihood and doesn't need the assumption of independence on variational posterior. On the other hand, MCMC could be used for full inference for kernel hyperparameters and avoids approximate marginal likelihood. Moreover, MCMC does not impose any requirement on the covariance function.
One of the most successful approaches to sparse GP is the framework of psudo-input GP in \citep{Snelson2005} which predicts the new test input as follows:
\begin{equation}
p(y|x,\Bar{X},\Bar{f})=N(y|k_{x}^{T}K_{M}^{-1}\Bar{f},K_{xx}-k_{x}^{T}K_{M}^{-1}k_{x}+\sigma^2)
\end{equation}
where $[K_M]_{mm'}=k(\Bar{x}_{m},\Bar{x}_{m'})$ and $[k_{x}]_{m}=k(\Bar{x}_{m},x)$
and the size of psudo-input is M which is much less than N(the size of original input)
Finding pesudo-input locations $\Bar{X}$ and hyperparameters is usuall done by maximizing the following marginal likelihood by gradient ascent.
\begin{equation}
p(y|X,\Bar{X},\theta)=\int p(y|X,\Bar{X},\Bar{f})p(\Bar{f}|\Bar{X}) d\Bar{f}
\end{equation}

One drawback of using pesudo-input sparse GP is that $MD+|\theta|$ parameters are needed to be fit instead of just $|\theta|$ which increases the risk of overfitting significantly.

\subsection{Infinite GMM}
Although GMM is extensively used for clustering such as \citep{He2011}\citep{Loffler2021},\citep{Liu2010},\citep{Muzeau2020}  and in most of these articles, expectation maximization is used significantly, but finding the right number of GM components from the data is still
a current topic of research. 

Infinite GMM is first described by \citep{Rasmussen1999}, and many ideas around it exists in the literature such as \citep{Tomoharu2013}.

\subsection{MCMC And Fast Convergence}
There are two major motivations for using MCMC for gaussian processes: \\
1- using MCMC as an alternative to maximum likelihood estimation(MLE) to estimate hyperparameters since MLE is just a point estimator. See \citep{Lindholm2015} for Monte Carlo approaximation \\
2- when the likelihood is nongaussian as in the case of time series models or when the noise is nongaussian. \\
The focus of the present paper is on hyper-parameter inference for Gaussian Processes. An MCMC method is good if it has shorter burn-in period, and uses a better proposal distribution. Laplace approximation is just one of the ways to obtain a good proposal distribution as is explained in \citep{Chowdhury2019}

\citep{Titsias2011} shows how to use MCMC for GP. In most practical cases in GP regression where likelihood is not gaussian, exact inference is intractable and therefore variational inference and MCMC are quite useful. For these reasons GP is generalized to consider non-gaussian noise as well as non-gaussian process as is explained in \citep{Snelson2003} by introducing warped GP which basically makes a transformation from the true observation space to the latent space. To find the distribution in the observation space, the Gaussian goes through the nonlinear warping function and the shape of the distribution may become asymmetric and multimodal.
Although deterministic approximate methods are very popular but they are limited to the case where likelihood is factorizable. MCMC is a good framework for non-deterministic approximate inference since in the long run it provides a precise estimate of the posterior and is not just a point estimator. Thus in general there are two major approaches to tackle non-gaussian likelihoods:  \\
1- Sampling methods that obtain samples of posterior \\
2- Approximation of posterior with some known form \\
The second approach needs good knowledge for the theory of markov chains in the general space. Since the seminal result of \citep{Mengersen1996} , convergence of MCMC has improved greatly. For example, in the framework of drift and lyapunov functions, \citep{Johndrow2018} has obtained a sharper bound for convergence, and \citep{Conrad2015} observed that when the likelihood has some local regularity, the number of model evaluations per MCMC step can be significantly reduced. MCMC methods such as block-based Metropolis-Hastings are much more efficient than Gibbs samplers and function variables are divided into disjoint sets corresponding to different function regions. So the proposal distribution is constructed by partitioning function values into some groups. The full details is explained in \citep{Lawrence2008} using some control variables which are basically some auxiliary function values which provide a low dimensional representation of the function.

\citep{Wang2013} has used MCMC based on the framework of "temporary mapping and caching" and implements the MCMC in two ways.

Simple approaches to hyper-parameter estimation of GPs like MLE or MAP understates the posterior uncertainty and therefore full bayesian approach is necessary as is mentioned in  \citep{Flaxman2015} and MCMC is used for inference for both posterior function and posterior over hyper-parameters.
Using Markov sampling methods for hyper-parameter estimation of GP is growing. see \citep{Inigo2015}.
\citep{Lalchand2020} has used hierarchical specification of GPs to deal with intractable hyper-parameter posteriors. The first approach they proposed is Hamiltonian Monte Carlo (HMC) and their second approach is based on approximation by a factorized gaussian and using variational inference.

\section{Posterior on function and hyper-parameters}
Unlike the classical point estimation of hyper-parameters using MLE or MAP, sampling methods is used in the present article. Using Bayes rule the posterior over function is 
\begin{equation}\label{posterior_over_function}
p(f|y)=\frac{p(y|f)p(f)}{\int p(y|f)p(f)df}
\end{equation}
Thus, for unseen inputs:
\begin{equation}\label{unseen_inputs}
p(f_{\star}|y)=\int p(f_{\star}|f)p(f|y)df
\end{equation}
where the conditional prior in \eqref{unseen_inputs}  is:
\begin{equation}
p(f_{\star}|f)= N(f_{\star}|K_{f_{\star},f} K_{f,f}^{-1}f, K_{f_{\star},f_{\star}}-K_{f_{\star},f} K_{f,f}^{-1}K_{f_{\star},f}^{T}      )
\end{equation}
Thus the prediction of values $y_{\star}$ is possible using:
\begin{equation}
p(y_{\star}|y)=\int p(y_{\star}|f_{\star})p(f_{\star}|y)df_{\star}
\end{equation}
when likelihood is gaussian, everything is tractable. The posterior GP depends on the value of kernel parameters $\theta$ as well as the likelihood parameters $\alpha$ which is related to noise level. Augmenting them together, $(\alpha),\theta$ are the hyper-parameters of the GP model. The idea is efficiently sampling from posterior $p(f|\alpha,\theta,y)$ which is a an extremely high dimensional random variable and therefore challenging. As mentioned earlier, there is wide spectrum of methods ranging from simple Gibbs sampling and Metropolis Hastings (MH) to modern methods such is block MH , control variables framework and finally the hardest one which is called full bayesian. MH is used in the present article.

\section{Metric embedding}
\citep{Ittai2011} covers some of the most important and practical approaches to metric embedding. The present paper is only focused on embedding into $\ell_{2}$ but the meta idea could be used for other methods as well.
Let $(x,d_{X})$ and $(Y,d_{Y})$ be metric spaces. An injective mapping $g:(X,d_{X}) \to (Y,d_{Y})$ is called D-embedding, where $D\geq 1$ is a real number, if there is a number $q>0$ such that for all $x,y\in X$, \\
\begin{equation}\label{distortion}
qd_{X}(x,y)\leq d_{Y}(g(x),g(y)) \leq Dqd_{X}(x,y)
\end{equation}
The distortion of g is the infimum of the numbers D such that g is a D-embedding. $d_{X}$ is the shortest path metric on graph while $d_{Y}$ is the metric on $\ell^{r}_{2}$.
The first step in Algorithm~\ref{alg:clustering1} is picking r vertices randomly. Therefore the distortion of the embedding is very dependent on this randomization step. There are many methods to embed into $\ell^{r}_{2}$ in the literature such as using the top eigenvectors of the graph laplacian. Although each embedding method may produce small or big distortion, the method of embedding in the present paper is motivated by \citep{Linial2001} and is similar to Frechet embedding which specifies a mapping $g_{i}:X \to \ell_2$
\begin{equation} \label{embedding}
g_{i}(x):=d_{X}(x_{i},x)
\end{equation}
where the final embedding  $ \oplus_{i=1}^{r} g_{i}$ is a map $g:X \to \ell_{2}^{r}$

\section{Learning induced conductance}
The approach of the present paper to graph partitioning is inspired by big progress in machine learning, and is therefore very different than methods of spectral graph theory or approximation algorithms.
Gaussian process is used to learn local graph conductance at any vertex.

\begin{figure}[H]
\centering
\includegraphics[scale=0.5]{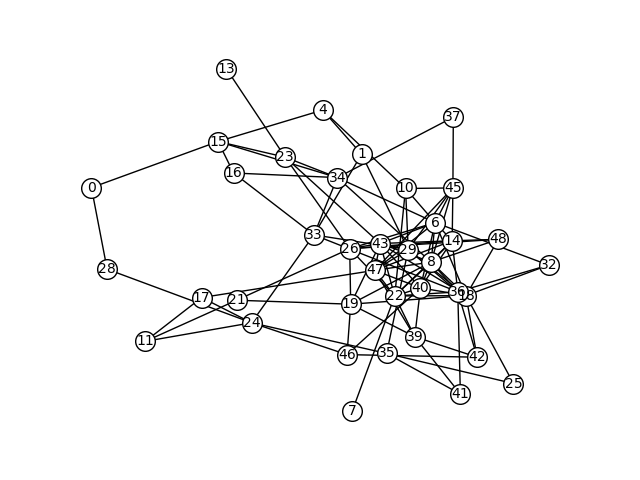}
\caption{50 vertices of MNIST dataset}\label{graph}
\end{figure}

Consider a Graph $G=(V,E)$, the ergodic flow of any two subsets of vertices $S_1$, $S_2$ is :
\begin{equation}
Q(S_{1},S_{2})=\sum_{i \in S_{1} , j \in S_{2}} Q_{ij}
\end{equation}
Conductance $\Phi$ of a subset of states of a Markov chain is \\
\begin{equation} \label{subsetconductance}
\Phi(S)=\frac{Q(S,\bar{S})}{\pi_{S}}=\frac{\underset{i\in S,j\in \bar{S}} {\sum} \pi_{i}p_{ij}}{\pi_S}
\end{equation}
where $\bar{S}$ in \eqref{subsetconductance} is $V-S$ and $\pi_{S}$ is the probability that Markov chain is at some state in S when stationarity is reached. The conductance of a Markov chain is the minimum conductance over all subsets S with $\pi(S)\leq 1/2$
\begin{equation}\label{conductanceofMC}
\Phi =   \underset{S\subset V, \pi_{S}\leq 1/2 } {min} \Phi(S)
\end{equation}
\eqref{conductanceofMC} is not flexible and is formulated as:
\begin{equation}
\Phi(r) =   \underset{S\subset V, \pi_{S}\leq r } {min} \Phi(S)
\end{equation}
From now on, conductance means conductance of a set. 
Finding the conductance in a static graph is easy while if the graph is dynamic like the evolution of communities in Twitter, the algorithm in \citep{Galhotra2015} could be used to find conductance. MNIST dataset contains 60000 training images with 28x28 pixels but 50 images is shown in Figure~\ref{graph} for illustration. 
A weighted graph is made as is depicted in Figure~\ref{graph} and the shortest path distance is used for any two vertices in this graph.
The weight is chosen equal to the euclidean distance between any two adjacent vertices.
The GP idea is that if two vertices are close to each other, their conductance should be similar.

\begin{definition}
uniform induced conductance of a vertex is defined as follows:
\begin{equation}\label{induced}
\Phi^{ind}_{r}(x) =   \underset{S\subset B(x,R):S=B(x,z),z\leq R, \pi_{S}\leq r } {min} \Phi(S)
\end{equation}
where $B(x,z)$ in \eqref{induced} is the ball of radius z
\end{definition}

Algorithm~\ref{alg:clustering1} shows how to use GP to learn uniform induced conductance and then infer the conductance for unseen datapoints without dealing with computational complexity of finding induced conductance for test data points.

\subsection{Learning by GP}
\begin{figure}[H]
\centering
\includegraphics[scale=0.5]{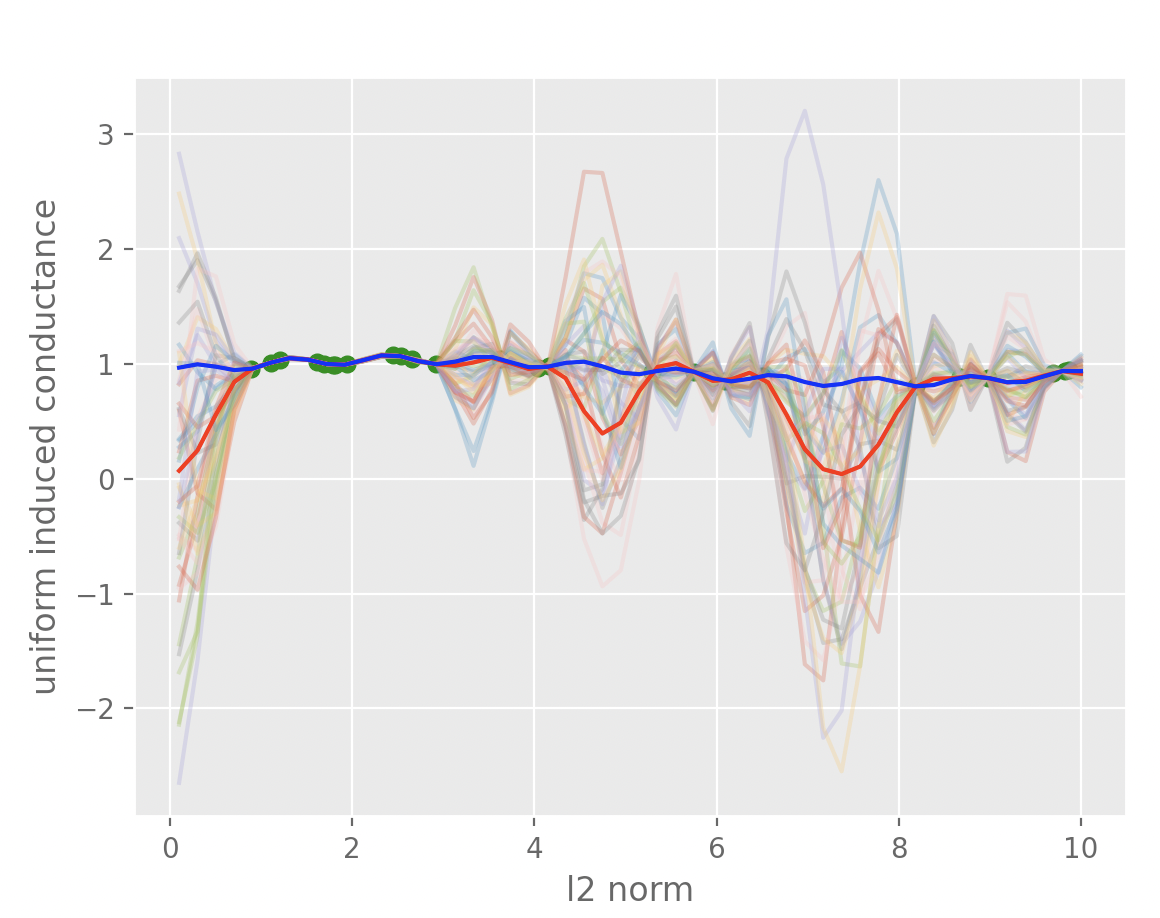}
\caption{induced conductance after GP learning in MNIST graph}\label{conductance}
\end{figure}

Figure~\ref{conductance} shows the samples that posterior has generated and in some regions uncertainty for induced conductance is high. The x-axis is the $\ell_{2}$ norm for vertices after metric embedding. Cholesky decomposition $K=LL^{T}$ is used for the implementation since usual inversion is prone to inaccuracy. The red line in figure~\ref{conductance}  is the mean $\mu=K_{\star}^{T}L^{-T}m$. The covariance is $\Sigma= K_{\star \star}-K^{T}_{\star}\beta$ where $\beta=(LL^{T})^{-1}K_{\star}$ .

\subsection{MCMC for hyperparameter estimation}
\begin{figure}[H]
\centering
\includegraphics[scale=0.5]{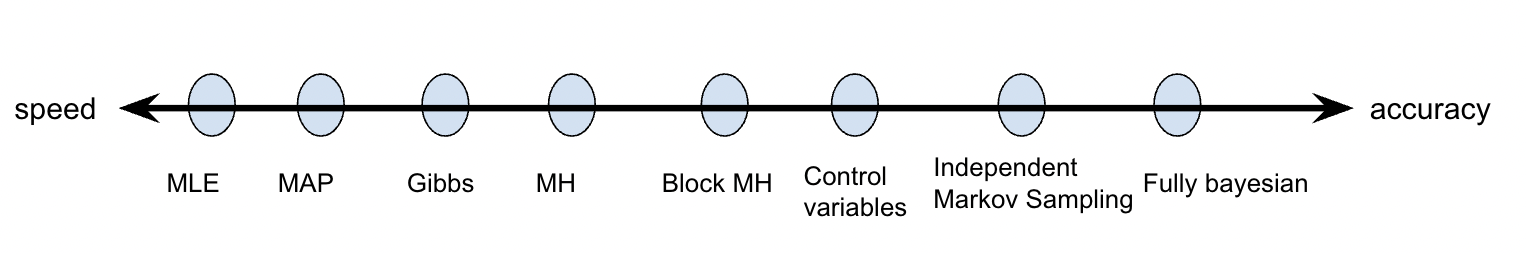}
\caption{spectrum of methods for hyper-parameter estimation of gaussian processes}\label{spectrum}
\end{figure}

In the theory of Markov chain in the general state space, methods like drift and small sets are very important to achieve fast convergence. The proposal distribution could be designed using these ideas to control mixing rate of these Markov chains.
There exists a spectrum of methods for hyper-parameter estimation as is shown in Figure~\ref{spectrum}. The quickest and most popular approaches are maximum likelihood estimation(MLE) and (Maximum a posteriori)MAP and are useful for point estimation only. To be more accurate and to measure uncertainty, methods like Gibbs sampling and MCMC methods like Metropolis Hastings is frequently used for measuring uncertainty of hyper-parameters. There is a tradeoff between accuracy and speed of the algorithm as is depicted in Figure~\ref{spectrum}. 
 \citep{Neal2000} developed a MCMC method for a software  to handle non-Gaussian noise and showed some examples for classification problems of relatively moderate size and used the following gamma prior for hyper-parameter $\phi=\theta^{-2}$ 
\begin{equation}\label{prior-parameter}
p(\phi)=\frac{(\alpha/2\omega)^{\alpha/2}}{\Gamma(\alpha/2)}\phi^{\alpha/2-1}\exp(-\phi \alpha/2\omega)
\end{equation}

Assume the hyper-parameters are ($\alpha$,$\theta$). In the fully bayesian framework, priors are assigned to these hyper-parameters and 
the conditional posterior distribution $p(\alpha,\theta|f,y)$ could be factorized across $\alpha,\theta$ as follows:
\begin{equation}\label{eq-posterior}
\begin{split}
p(\alpha,\theta | f,y)=p(\alpha | f,y) p(\theta | f)   \\
p(\alpha | f,y) \propto p(y|f,\alpha)p(\alpha) \\
p(\theta | f) \propto p(f|\theta)p(\theta)
\end{split}
\end{equation} 
Assume the proposal distribution is a Gaussian, then the MH algorithm can be applied to the posterior in \ref{eq-posterior}.
The posterior distribution of the hyper-parameters and latent variables is achieved:
\begin{equation}
p(f_{\star},f,\theta | y)=\frac{p(y|f)p(f_{\star},f |\theta)p(\theta)}{p(y)}
\end{equation}
Thus, the predictive distribution of $f_{\star}$ is:
\begin{equation}\label{predictive}
\begin{split}
p(f_{\star}|y)&=\int p(f_{\star},f,\alpha,\theta|y)df d\alpha d\theta   \\
&=\frac{1}{p(y)}\int p(y|f)p(f_{\star},f|\alpha,\theta)p(\alpha)p(\theta)dfd\alpha d\theta
\end{split}
\end{equation}

It is known that if noise is gaussian, the result is analytically tractable. However, even if the noise is gaussian, the integral over $\alpha,\theta$ (such as the case of using a type of prior in \ref{prior-parameter}) is intractable in general and therefore MCMC could be used to perform this integration.

\begin{algorithm}[H]
\caption{outputs clusters }
\label{alg:clustering1}
Input: A weighted undirected graph with n vertices and m edges \\
loop: \\
1: pick r reference vertices randomly \\
2: create a data structure by metric embedding into $\ell^{r}_{2}$ using \eqref{embedding} \\
3: pick some training points and calculate induced conductance using \eqref{induced} \\
4: learn the gaussian process using training pairs (embedding coordinate, induced conductance) and use MCMC method to estimate
hyperparameters robustly. \\
5: pick some test points and isolate balls with high induced conductance and make cluster \\
Output: clusters 
\end{algorithm}

\section{Conclusion}
 Uniform induced conductance is defined as a measure of expansion around each vertex. Learning uniform induced conductance provides approximate qualitative bahaviour  around each vertex of a graph which could be seen as a method for graph partitioning using gaussian processes. 
\bibliographystyle{agsm}
\bibliography{bayesian}
\end{document}